%% file: main.tex
\documentclass[10pt,journal,compsoc]{IEEEtran}
\newif\ifpeerreview

\peerreviewfalse

\usepackage[nocompress]{cite}
\usepackage{url}
\usepackage{amsmath,amssymb,graphicx}

\usepackage{lipsum} 

\usepackage{diagbox} 
\usepackage{algorithm}
\usepackage{algpseudocode}
\usepackage{array}
\usepackage{arydshln} 
\usepackage{xcolor}
\usepackage{hyperref}

\usepackage[switch]{lineno}

\newcommand{\paperID}{24}

\title{\name{}: Drone and Ground Gaussian Splatting for 3D Building Reconstruction}

\author{Yujin Ham, 
        ~Mateusz~Michalkiewicz,
        ~Guha~Balakrishnan 
\IEEEcompsocitemizethanks{\IEEEcompsocthanksitem Y. Ham, M. Michalkiewicz, and G. Balakrishnan are with the Department
of Electrical and Computer Engineering, Rice University, Houston, TX, 77005. E-mail: guha@rice.edu\protect\\
}
}

\begin{document}

\include{format_and_definitions}

\input{0-abstract}

\maketitle
\thispagestyle{empty}

\input{1-introduction}

\input{2-related}

\input{3-dataset}

\input{4-methods}

\input{5-experiments}

\input{6-conclusion}


\section{Acknowledgments}
This research is based upon work supported by the Office of the Director of National Intelligence (ODNI), Intelligence Advanced Research Projects Activity (IARPA), via IARPA R\&D Contract No. 140D0423C0076. The views and conclusions contained herein are those of the authors and should not be interpreted as necessarily representing the official policies or endorsements, either expressed or implied, of the ODNI, IARPA, or the U.S. Government. The U.S. Government is authorized to reproduce and distribute reprints for Governmental purposes notwithstanding any copyright annotation thereon. We thank our colleagues Ashok Veeraraghavan, Boyang (Tony) Yu, and Cheng Peng who provided who provided helpful feedback and suggestions.

\bibliographystyle{IEEEtran}
\bibliography{references}

\ifpeerreview \else


\begin{IEEEbiography}
[{\includegraphics[width=1in,height=1.25in,clip,keepaspectratio]{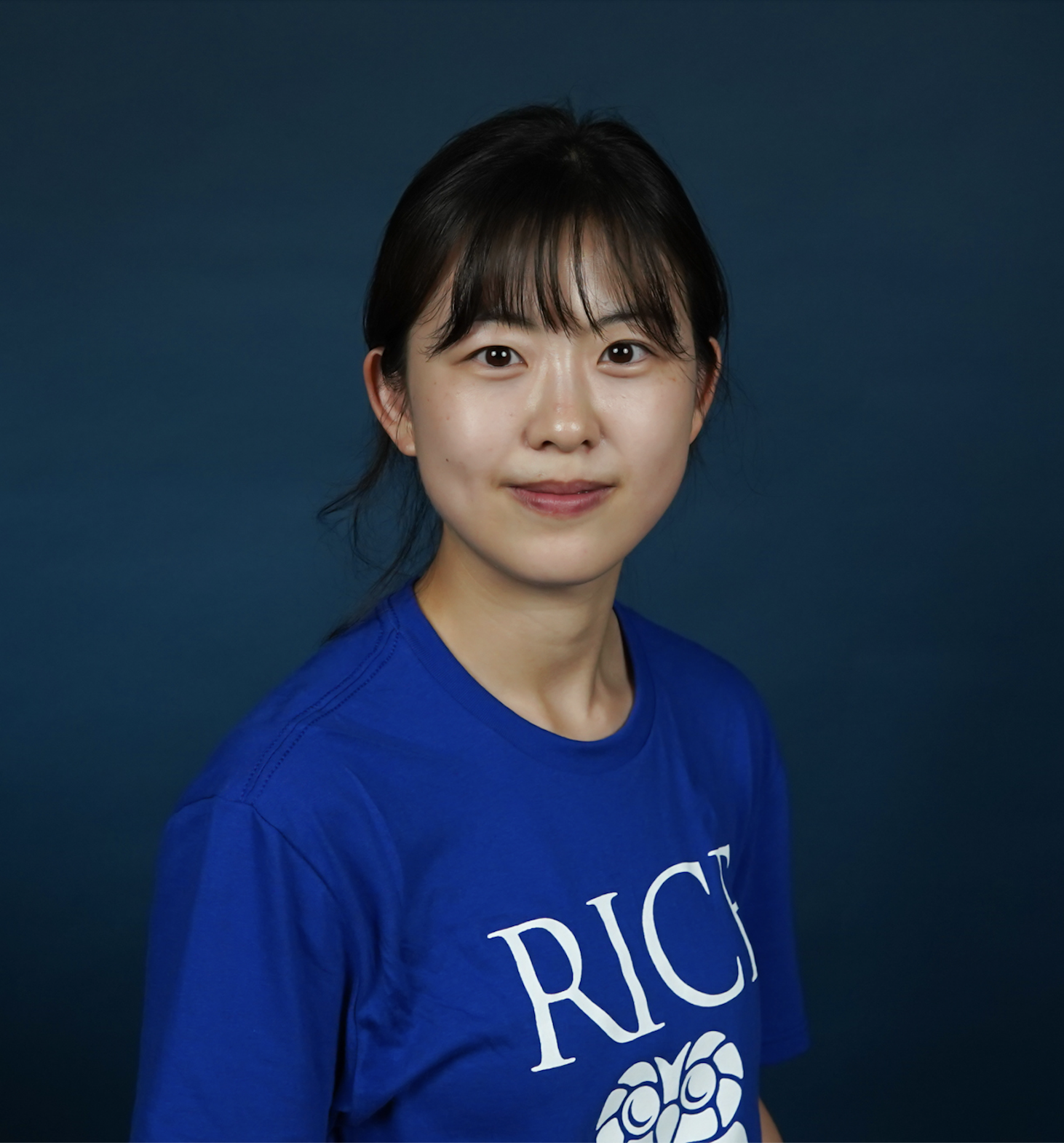}}]{Yujin Ham}
is a Ph.D. candidate in the Department of Electrical and Computer Engineering at Rice University, Houston, TX. She received her M.S. in 2022 and B.S. in 2020 in the department of Electronics and Electrical Engineering at Ewha Womans University, Seoul, South Korea. Her research focus lies in computer vision and 3D scene understanding.
\end{IEEEbiography}

\begin{IEEEbiography}
[{\includegraphics[width=1in,height=1.42in,clip,keepaspectratio]{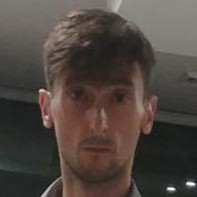}}]{Mateusz Michalkiewicz}
is a Postdoctoral Associate in the Department of Electrical and Computer Engineering at Rice University in Houston, US. He received his Ph.D. degree from the University of Queensland in Brisbane, Australia, where he was supervised by Anders Eriksson and Mahsa Baktashmotlagh. His current research interests lie in computer vision and vision and language models.
\end{IEEEbiography}

\begin{IEEEbiography}
[{\includegraphics[width=1in,height=1.42in,clip,keepaspectratio]{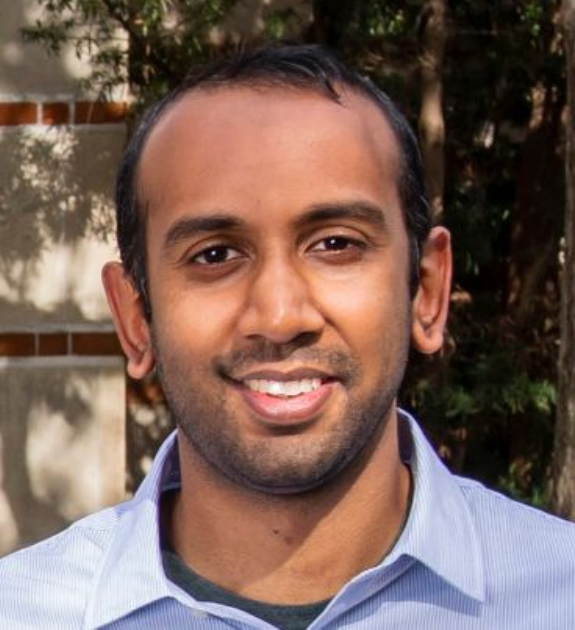}}]{Guha Balakrishnan}
is an Assistant Professor in the Electrical and Computer Engineering (ECE) department at Rice University, where he leads the Rice Visual Intelligence Group (RVIG). He completed his Ph.D. in 2018 in MIT's EECS department. He works in the fields of computer vision and graphics, particularly on problems in algorithmic fairness, 3D reconstruction, and clinical applications.
\end{IEEEbiography}

\fi

\end{document}

%% file: format_and_definitions.tex
\newcommand{\mm}[1]{\textcolor{blue}{\bf [MM: #1]}}

\newcommand{\gb}[1]{\textcolor{olive}{\bfseries [GB: #1]}}

\newcommand{\yh}[1]{\textcolor{orange}{\bf [YH: #1]}}

\newcommand{\name}{DRAGON}
\newcommand{\dsname}{Buildings-NVS}

\newcommand{\bsname}{BASIC}

\newcolumntype{C}[1]{>{\centering\arraybackslash}m{#1}}

\newcommand{\ie}{\textit{i.e.}\ }

%% file: 0-abstract.tex
\IEEEtitleabstractindextext{%
\begin{abstract}
3D building reconstruction from imaging data is an important task for many applications ranging from urban planning to reconnaissance. Modern Novel View synthesis (NVS) methods like NeRF and Gaussian Splatting offer powerful techniques for developing 3D models from natural 2D imagery in an unsupervised fashion. These algorithms generally require input training views surrounding the scene of interest, which, in the case of large buildings, is typically not available across all camera elevations. In particular, the most readily available camera viewpoints at scale across most buildings are at near-ground (e.g., with mobile phones) and aerial (drones) elevations. However, due to the significant difference in viewpoint between drone and ground image sets, camera registration -- a necessary step for NVS algorithms -- fails. In this work we propose a method, \name{}, that can take drone and ground building imagery as input and produce a 3D NVS model. The key insight of \name{} is that intermediate elevation imagery may be extrapolated by an NVS algorithm itself in an iterative procedure with perceptual regularization, thereby bridging the visual feature gap between the two elevations and enabling registration. We compiled a semi-synthetic dataset of 9 large building scenes using Google Earth Studio, and quantitatively and qualitatively demonstrate that \name{} can generate compelling renderings on this dataset compared to baseline strategies. Our dataset and results are available \href{https://yujinh22.github.io/publication/dragon}{on the project webpage}.

\end{abstract}

\begin{IEEEkeywords} 
3D Building Reconstruction, \and Novel View Synthesis, \and 3D Gaussian Splatting, \and Multi-Elevation Reconstruction \end{IEEEkeywords}
}

%% file: 1-introduction.tex
\IEEEraisesectionheading{
  \section{Introduction}\label{sec:introduction}
}
\IEEEPARstart{3}{d} building reconstruction is an important task that is useful in a variety of applications from urban planning and monitoring to disaster relief and reconnaissance. Traditional building reconstruction methods assume a data modality such as aerial LiDAR~\cite{chen2005building,elaksher2002reconstructing,forlani2003building} as input, which is expensive and offers limited viewpoint coverage. The proliferation of standard imaging sensors, along with powerful recent advances in Novel View Synthesis (NVS) algorithms like Neural Radiance Fields (NeRF)~\cite{mildenhall2021nerf} and 3D Gaussian Splatting (3DGS)~\cite{kerbl20233d}, offer the potential to develop 3D building models in a cheap and scalable manner. Natural images of buildings are most readily acquired at two elevations: near-ground (such as those taken by mobile phones) and aerial (such as those taken by drones), representing highly contrasting structural viewpoints (see Fig.~\ref{fig:teaser}). 3D reconstruction that can work with such sparse footage without needing intermediate elevation imagery would enable large-scale structural modeling for a variety of otherwise inaccessible scenarios. For this reason, in this work, we aim to develop a NVS-based method that can combine \emph{only} drone and ground building imagery into a coherent 3D model of a building's outer structure.

\input{figures_tex/ds_samples}

Standard variants of NeRF and 3DGS algorithms assume a set of input 2D images covering a continuous range of viewpoints around a scene. These algorithms first register the views together into a common reference system by finding matching features, and then combine their information into a 3D model via a rendering-based optimization. 

Unfortunately, the crucial registration step poses significant challenges when given only aerial and ground footage as input. As shown in the examples in Fig.~\ref{fig:teaser}, the same structure on a building can look completely different in scale and viewpoint across elevations. Indeed, popular registration packages struggle on such data because of a lack of robust feature point matches. Hence, while aerial and ground building footage are widely accessible, they are apparently not immediately usable for 3D building reconstruction without further camera pose input metadata.


In this work, we address the challenge of large-scale building reconstruction where only drone and near-ground image sets are available, \emph{without} known camera poses. To do this, we draw on three insights. First, while registration methods fail given only drone and ground images, they are successful when intermediate elevation images are available. Second, view rendering methods such as NeRF and 3DGS are capable of limited \emph{extrapolation} beyond their training view range, likely due to inherent implicit regularization in their representations. Third, as shown in previous studies this extrapolation ability may be enhanced using perceptual regularization. We use these ideas to develop a framework, \name{} (for DRone And GrOuNd Rendering). \name{} begins by training a 3D model on aerial footage only, providing an initial rendering model with full context of the coarse layout of the structure. The algorithm then iteratively alternates between generating progressively lower-elevation imagery and updating its representation with previously generated views. \name{} eventually reaches the lowest elevation (ground), at which point all of the real and generated views are used to perform registration and generate a final model. \name{} is conceptually simple, and requires no further annotation or information per view to operate. 

\input{tables/tab_buildings_dataset}

We demonstrate the effectiveness of \name{} using the 3DGS rendering algorithm on a new semi-synthetic dataset we built called \textit{\dsname{}}, consisting of imagery of 9 building sites from Google Earth Studio (see Table~\ref{tab:scenes} and Fig.~\ref{fig:ds_samples}). Results first show that popular registration packages like COLMAP~\cite{schoenberger2016sfm} fail to register the drone and ground footage on any of these scenes. However, \name{} aligns 97.47\% of the input views (with average 0.01m position and 0.12$^{\circ}$ rotation errors). With camera poses obtained using \name{}, we demonstrate that \name{} produces high-quality renderings across all viewing angle/elevation scenarios. Finally, we show that adding perceptual regularization encoded by deep feature spaces like DreamSim~\cite{fu2023dreamsim} and OpenCLIP~\cite{cherti2023reproducible} further improves quantitative and qualitative performance. We conclude by discussing limitations and considerations when using \name{}.

%% file: figures_tex/ds_samples.tex
\begin{figure}[t!]
\includegraphics[width=\linewidth]{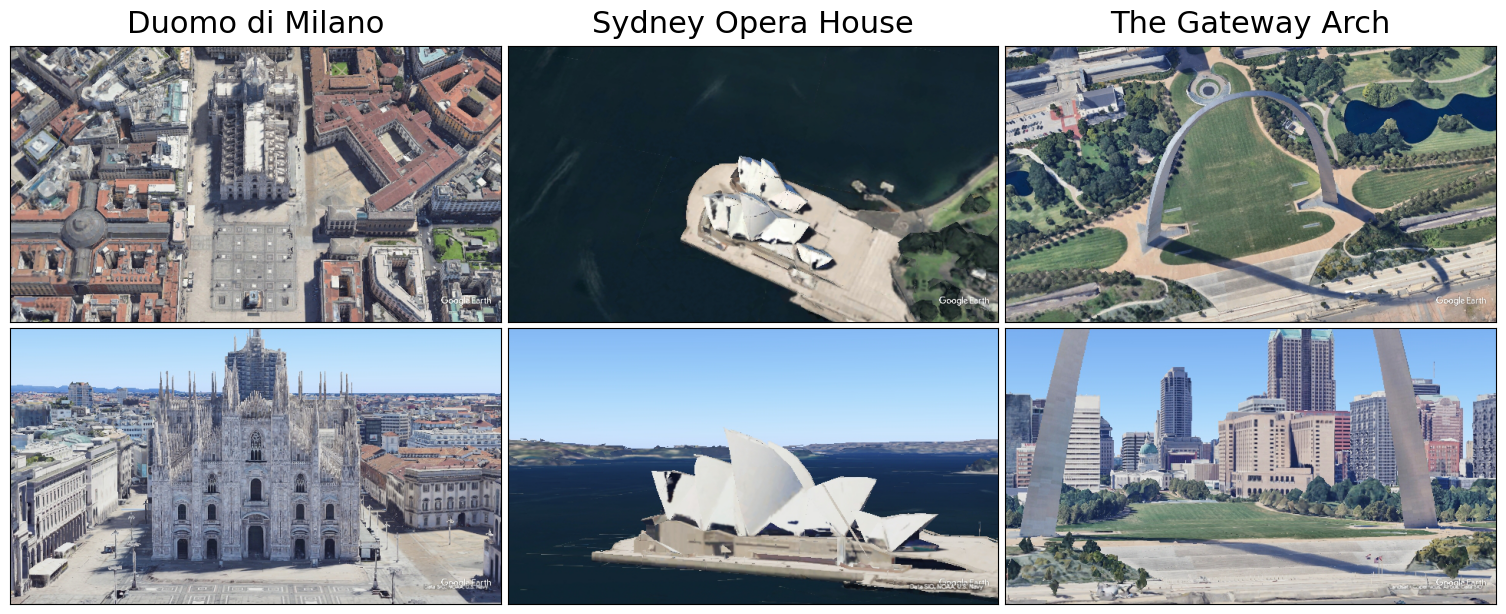}
\caption{\textbf{Building-NVS, a dataset of images we collected from Google Earth Studio.} We show views of three large buildings from drone-level (top) and ground-level (bottom) elevations. These opposing elevations offer highly contrasting viewpoints of the physical structures, which can provide complementary information towards 3D modeling. However, the lack of easily matching visual features across elevations inhibits registration, a key step in novel view synthesis algorithms like NeRF~\cite{mildenhall2020nerf} and 3D Gaussian Splatting~\cite{kerbl20233d}.}
\label{fig:teaser}
\end{figure}

%% file: tables/tab_buildings_dataset.tex
\begin{table*}[t!]
    \caption{\textbf{Camera altitudes/trajectory radii/target altitudes (all measured in meters) per elevation for each building in our collected dataset \dsname{}.} The targeted altitude refers to the point where the perpendicular line from the center of the building intersects the plane at the targeted height where the camera is aimed. Each orbit from 5 different elevations consists of 61 images positioned with evenly spaced camera locations. We selected these values manually by focusing on relevant perceptual characteristics while keeping most of the structure in view.
    }
    \resizebox{\linewidth}{!}{
        \begin{tabular}{|C{2cm}|*{9}{C{1.5cm}|}}
        \hline
           {\diagbox[width=2.4cm]{Elevation}{Building}} &  Arc de Triomphe & Colosseum & Duomo di Milano & Eiffel Tower & Himeji Castle & Piazza del Duomo & Space Needle  & Sydney Opera House & The Gateway Arch \\
            \hline
           Ground & 101/250/89 & 77/313/52 & 177/250/153 & 101/438/89 & 92/250/67  & 58/250/34 & 83/188/71 & 60/438/48 & 166/313/166 \\
           Mid 1  & 201/313/89 & 127/375/52 & 250/438/152 & 201/500/89 & 154/313/67 & 146/250/34 & 195/250/71 & 148/438/48 & 240/375/166 \\
           Mid 2  & 301/375/89 & 252/438/52 & 402/500/177 & 351/625/114 & 279/375/67 & 246/313/34 & 345/250/95 & 298/563/48 & 403/438/203 \\
           Mid 3  & 451/500/89 & 452/500/52 & 600/438/152 & 601/750/151 & 404/375/67 & 346/313/34 & 495/375/120 & 498/563/48 & 603/563/203 \\
           Drone  & 701/313/89 & 702/500/52 & 802/375/152 & 1000/375/177 & 604/313/67 & 596/313/34 & 695/313/145 & 748/500/48 & 803/375/203 \\
           \hline
        \end{tabular}
    }
    \label{tab:scenes}
\end{table*}


%% file: 2-related.tex
\section{Related Work}
\label{sec:related}

\subsection{Building Reconstruction from LIDAR and Aerial Footage}
Several prior works estimate building structures from LIDAR footage~\cite{chen2005building,elaksher2002reconstructing,forlani2003building,huang2022city3d, wang20133d, zheng2017hybrid}. LIDAR directly provides 3D scene information, but is also expensive to acquire and therefore limited in availability. Several methods also try to infer 3D information of buildings from monocular (aerial) imagery alone~\cite{chen2021mask, li20213d, xu20223d}. Limited by the information from a single elevation, these methods offer coarse 3D details, such as building height and general shape. In contrast, we attempt to build full 3D rendering of outer building structures based on both aerial and ground images, without additional modalities. In addition, our technique builds on differentiable view rendering techniques like 3DGS, which is fundamentally different from these past approaches. 

\subsection{Neural/Differentiable Rendering}
Neural Radiance Fields (NeRF)~\cite{mildenhall2020nerf, Barron_2021_ICCV,muller2022instant} are a family of deep neural network techniques that have brought a new wave of ideas and state-of-the-art results to view synthesis. NeRF reconstructs scene parameters including geometry, scattered radiance, and camera parameters~\cite{lin2021barf} in an end-to-end fashion based on a simple reconstruction loss over the input views. NeRF uses neural networks to fit the scene parameters, which are capable of regressing on highly nonlinear, complex functions typical of real-world scenes. This leads to a continuous scene representation, which can be rendered immediately for any desired novel view simply by evaluating the representation for each pixel. 

Recently, 3D Gaussian Splatting (3DGS)~\cite{kerbl20233d} has emerged as a compelling alternative to NeRF. 3DGS explicitly represents a scene with (millions of) 3D Gaussian objects with optimizable 3D location and appearance parameters. Crucially, these parameters are optimized in an end-to-end fashion using a reconstruction loss by ``splatting'' the Gaussians onto 2D planes. 3DGS has proven far faster to train for large scenes than NeRF. Due to this reason, we use 3DGS in our experiments.

\subsection{Neural Rendering for Large Scenes}
NeRF and 3DGS also have some demonstrated successes on large scenes. The first work in this space was BlockNeRF~\cite{tancik2022block}, which modeled city blocks from ground-level footage. BungeeNeRF~\cite{xiangli2022bungeenerf} is the closest related study to ours, which reconstructs buildings using dense multi-elevation imagery from drone to ground elevations. That study pointed out that scale differences resulting from cameras positioned at varying altitudes can pose significant challenges to rendering. For example, the level of detail and spatial coverage varies significantly across different altitudes. These scaling problems are somewhat mitigated with follow-up NeRF variants~\cite{Li_2023_ICCV, turki2022mega, lin2024vastgaussian, tancik2022block}. Recently, VastGaussian~\cite{lin2024vastgaussian} extended 3DGS to handle large building scenes, mostly from aerial imagery. Unlike our problem setup, all of these algorithms assume a densely sampled set of input 2D views at orientations and elevations, and can not handle the case of combining only drone and ground footage into one coherent 3D model.

\input{tables/tab_intro_spsg}
\subsection{Registration and Camera Pose Estimation}
A crucial first step for NVS algorithms is image registration, i.e., obtaining the camera poses for each input image. This is commonly achieved by running the Structure-from-Motion (SfM)~\cite{schonberger2016structure} library COLMAP~\cite{schoenberger2016sfm}, designed for 3D reconstruction from 2D images. The process begins by extracting features from each image, followed by matching these features across multiple images to establish corresponding points~\cite{schoenberger2016sfm}. 
The output of registration includes intrinsic camera parameters, camera poses (extrinsic parameters), and 3D point clouds. Other image registration and camera pose estimation algorithms also exist, such as SLAM-based methods~\cite{engel2014lsd}, and deep learning models such as SuperGlue~\cite{sarlin2020superglue}. We empirically find that both COLMAP and SuperGlue using SuperPoint~\cite{detone2018superpoint} keypoints and descriptors (SPSG) struggle on our problem setup (see Table~\ref{table:intro_regi}). Some NeRF and 3DGS variants relieve the reliance on SfM prepocessing by incorporating camera pose estimation directly into the  optimization framework~\cite{fu2023colmap, wang2021nerf, lin2021barf}. 
These methods work reasonably for normal-sized objects and with sparse viewpoints that cover the full range of the scene, but do not work with large ranges of missing viewpoints, as in our problem setup. A recent study uses deep generative diffusion models~\cite{wang2023posediffusion} to aid in camera pose estimation, but was not demonstrated on extremely sparse viewpoints of large, multi-elevation scenes.




%% file: tables/tab_intro_spsg.tex
\begin{table*}[h]
\centering
\caption{\textbf{Popular registration packages struggle to register only drone and ground imagery.} This comparison evaluates the registration performance using COLMAP~\cite{schoenberger2016sfm} and SPSG~\cite{sarlin2020superglue, detone2018superpoint} on three scenes from our \textit{\dsname{}} dataset. SPSG exhibits superior registration results on average, however, its performance varies significantly depending on the building. For example, in the case of (b) Arc de Triomphe and (c) Piazza de Duomo, both drone and ground image sets can be successfully registered, though this comes with a noticeable increase in errors among the matched images. Conversely, for the (a) Eiffel Tower, SPSG is still unable to register both drone and ground images simultaneously and results in higher errors compared to COLMAP.} 
\label{table:intro_regi}
\resizebox{0.9\linewidth}{!}{
\begin{tabular}{|cc|cc|cc|cc|cc|}
\hline
\multicolumn{2}{|c|}{Building}                          & \multicolumn{2}{c|}{(a) Eiffel tower}  & \multicolumn{2}{c|}{(b) Arc de Triomphe} & \multicolumn{2}{c|}{(c) Piazza de Duomo}\\ \hline
\multicolumn{2}{|c|}{Method}                        & \multicolumn{1}{c|}{COLMAP} & SPSG & \multicolumn{1}{c|}{COLMAP} & SPSG  & \multicolumn{1}{c|}{COLMAP} & SPSG\\ \hline
\multicolumn{2}{|c|}{Matched (\%) $\uparrow$}       & \multicolumn{1}{c|}{50.00} & 50.00 & \multicolumn{1}{c|}{50.00}  & 100.00 & \multicolumn{1}{c|}{50.00}  & 100.00\\ \hline
\multicolumn{1}{|c|}{Errors for matched} & rotation ($^{\circ}$)        & \multicolumn{1}{c|}{0.31/0.08} & 0.53/0.09 & \multicolumn{1}{c|}{0.18/0.09} & 0.43/0.14 & \multicolumn{1}{c|}{1.15/0.55} & 2.01/0.12 \\ \hline
\multicolumn{1}{|c|}{(Avg/Std)}          & position (m) $\downarrow$    & \multicolumn{1}{c|}{0.01/0.01} & 0.03/0.02 & \multicolumn{1}{c|}{0.01/0.01} & 0.06/0.05 & \multicolumn{1}{c|}{0.06/0.04} & 0.15/0.07 \\ \hline
\end{tabular}
}
\end{table*}

%% file: 3-dataset.tex
\section{\dsname{} Dataset}
\label{sec:dataset}
\input{figures_tex/ds_overview}

\input{figures_tex/ds_coverage}

We introduce \textit{\dsname{}}, a dataset comprising multi-elevation imagery from 9 building sites using Google Earth Studio \footnote{https://www.google.com/earth/studio/} (see Table~\ref{tab:scenes} and Fig.~\ref{fig:ds_samples}). 
The 9 buildings are diverse in many aspects, including location (spanning four continents), height, architectural styles, backgrounds (e.g., water bodies, trees, other buildings), symmetry, and occlusions. For each building, we captured images over five camera elevations ranging from near-ground to high aerial elevations. For each elevation, we sampled 61 images evenly along a circle, with the camera pointed at a particular target 3D location which affects the vertical rotation of the camera. We manually selected the camera elevations, target elevations, and trajectory radii per building to focus on relevant perceptual characteristics while keeping most of the structure in view. We present these parameters in Table~\ref{tab:scenes}. Google Earth Studio permits a different lower limit on camera elevation per location, leading to different ground elevations across buildings in our dataset. We partitioned these images into training and testing subsets by elevation (see Fig.~\ref{fig:ds_coverage}. The training dataset consists of elevations $X^0$ (ground) and $X^4$ (drone), and the testing dataset comprises all elevations ($X^0$ to $X^4$). Each building has 122 training images and 305 testing images.


%% file: figures_tex/ds_overview.tex
\begin{figure}[t!]
\centering
\includegraphics[width=\linewidth]{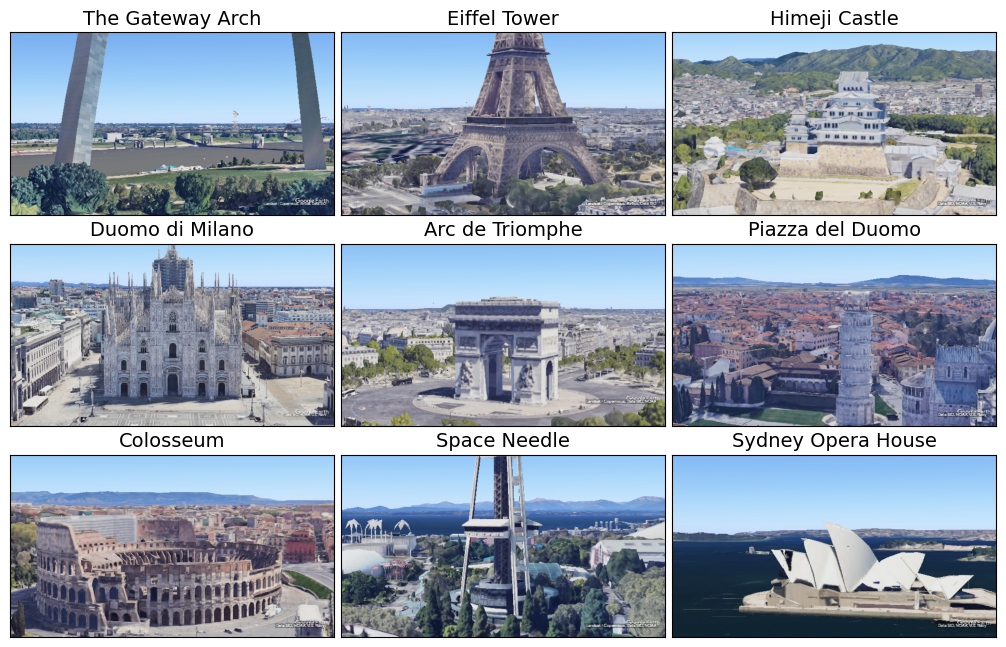}
\caption{\textbf{Visual depictions of the 9 buildings in our \dsname{} dataset.} These buildings have unique characteristics from height to architecture style and backgrounds.}
\label{fig:ds_samples}
\end{figure}

%% file: figures_tex/ds_coverage.tex
\begin{figure*}[t!]
\includegraphics[width=\textwidth]{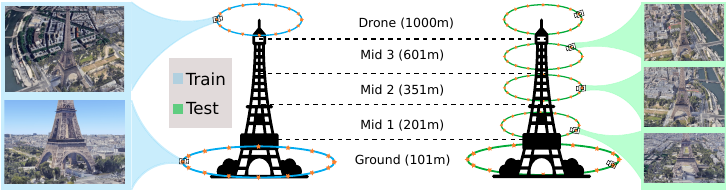}
\caption{\textbf{Train/test split of images for a given scene (Eiffel Tower) from our proposed \dsname{} dataset.} (Left) Training data covers only drone and ground elevations. (Right) Testing data covers all five elevations including drone and ground (due to space constrains, we provide sample images from middle three elevations in this figure).}
\label{fig:ds_coverage}
\end{figure*}


%% file: 4-methods.tex
\section{Methods}
\input{figures_tex/reg_ours}
\subsection{Background on 3D Gaussian Splatting (3DGS)}
\label{sec:3dgs_basic}
In 3D Gaussian splatting, a set of 3D Gaussian objects is used to represent a scene. Initialized with point clouds from structure-from-motion~\cite{schonberger2016structure}, each Gaussian is defined spatially by:
\begin{equation}
    G(x) = \exp\left(-\frac{1}{2}(x - \mu)^\top\Sigma^{-1}(x - \mu)\right),
\end{equation}
where $\mu$ and $\Sigma$ represent the spatial average and the covariance matrix. Additionally, each Gaussian has an opacity factor $o$ and a color $c$ conditioned on viewing perspective. During rasterization, each 3D Gaussian is projected onto the 2D image space from a particular viewing angle via a projection matrix. 
The color of a pixel in a rendered image is obtained by alpha-blending $N$ 2D Gaussians arranged in a depth-aware sequence from the nearest to the farthest relative to the camera viewpoint's perspective:
\begin{equation}
    C = \sum_{i \in N} \tau_i\alpha_i c_i \quad \text{with} \quad \tau_i = \prod_{j=1}^{i-1} (1 - \alpha_j),
    \label{eq:3dgs}
\end{equation}
where $\alpha$ is a product of the opacity $o$ and the likelihood of the pixel coordinate in image space. 
The Gaussian parameters are then optimized using a rendering loss function over the input 2D views:
\begin{align}
    L_{3DGS}(I, \hat{I}) &=  L_1(I, \hat{I}) + \lambda_{ssim} L_\text{ssim}(I, \hat{I}), \label{eq:3dgs_loss}
\end{align}
where $L_1(\cdot, \cdot)$ measures pixelwise L1 distance, $L_\text{ssim}(\cdot, \cdot)$ measures SSIM~\cite{wang2004image} distance, and $\lambda_{ssim}$ is a tradeoff hyperparameter.

\subsection{\name{}}
We assume as input two sets of images $X_{ground}$ and $X_{drone}$, depicting the building from ground and drone elevations (see Fig.~\ref{fig:ds_coverage}). We assume these images are reasonably distributed at different viewing angles in order to give a full 360-degree coverage of the building. Our goal is to develop one 3D NVS algorithm (3DGS in our experiments) based on this imagery.

An obvious first approach to solve this problem is to simply run the NVS model on the provided views, described in Alg.~\ref{alg:basic} (BASIC). However, this requires camera poses as input, produced by a registration package such as COLMAP~\cite{schoenberger2016sfm}. Such registration algorithms typically use Structure-from-Motion techniques~\cite{haming2010structure}, using a key assumption that the input set consists of overlapping images of the same object, taken from different viewpoints. However, in the case where only drone and near-ground image sets are available, visual overlap across sets is limited. Furthermore, even if there is visual overlap between image sets, high disparity in feature quality makes it challenging to find correspondences.

We do find, however, that registration succeeds when intermediate level images are provided, serving as a ``bridge'' between the ground and aerial elevations (see Fig.~\ref{fig:teaser}). This insight motivates \name{}, described in Alg.~\ref{alg:dragon}, which extrapolates intermediate elevation imagery from the drone altitude views in an iterative manner (see Sec.~\ref{subsec:iterative_scheme}). After generating images across all elevations, \name{} registers all views (real and generated) together (lines 7-8 of Alg.~\ref{alg:dragon}). Finally, using the original training views only along with their (now inferred) camera poses, we construct a 3D model using an NVS algorithm like 3DGS (line 9). We describe details of the iterative process and loss function in the following sections. 

\input{tables/tab_notation}
\input{algorithms/baseline_alg}
\input{algorithms/dragon_alg}

\input{figures_tex/perceptual_loss}


\subsubsection{Intermediate Elevation Image Generation}
\label{subsec:iterative_scheme}

We generate intermediate elevation images in an iterative manner, consisting of repeated steps of registration and extrapolated view rendering. We begin the process using only drone input images because they are easier to register to one another with high accuracy (demonstrated in our experiments, see Table~\ref{table:registration}). Two reasons for this are that ground elevations exhibit occlusions caused by neighboring buildings, and that drone images have more overlap with one another in content, resulting in easier feature matching. As detailed in lines 3-5 of Alg.~\ref{alg:dragon}, we iteratively expand the viewing range of the NVS model by registering the current cumulative input image set, retraining the NVS model using these images and registered poses, generating new (extrapolated) views at camera poses in the next lowest elevation level, and adding these new views to the cumulative image set. Eventually, this process reaches the ground elevation, and stops. 

\subsubsection{Perceptual Regularization for Improved Extrapolation} 
\label{subsec:perceptual}

Novel view synthesis methods like 3DGS work best when \emph{interpolating} novel views that are nearby a subset of input training views. Conversely, 3DGS performance degrades rapidly when \emph{extrapolating} target views that are significantly far from training viewpoints, as illustrated in Fig.~\ref{fig:vanilla}. Rendering intermediate elevation imagery from drone and ground imagery is closer to extrapolation since the input views are so far apart, posing a significant challenge. Rendering errors affect both registration (propagating errors in landmarks), and view synthesis steps of \name{} (see Fig.~\ref{fig:result-rendering-b1}).

To address this challenge, we propose adding perceptual regularization to the basic 3DGS loss function in Eq.~\ref{eq:3dgs_loss} to encourage extrapolated rendered views to be perceptually similar to nearby 3DGS training views. Perceptual similarity metrics have been well-studied in computer vision, and popular current metrics are based on measuring distances in feature spaces encoded by deep neural networks. We consider two such models, DreamSim~\cite{fu2023dreamsim} and OpenCLIP~\cite{cherti2023reproducible}. DreamSim encodes features particularly designed to correlate with humans on semantic visual perception tasks. We find in our own analysis (see Fig.~\ref{fig:dreamsim}) that DreamSim is superior to the widely used LPIPS metric for our task. LPIPS tends to perceive significant differences between image patches that have some geometrical distortions, even when they share the same semantic content. This is a negative property for our task, since we intend to compare images with slight changes to camera pose. In contrast, DreamSim prioritizes perceptual image similarity over geometric disparities, rendering it more suitable as an auxiliary loss function for \name{}. We additionally consider OpenCLIP because it captures a different type of semantically-guided representation to DreamSim based on language. Such vision-language models have been successfully shown to provide regularization for NVS methods, including NeRF-based approaches \cite{jain2021putting, wang2022clip, feng2022viinter, lee2024posediff} as well as diffusion-based ones \cite{deng2023nerdi, jain2022zero, liu2023zero}.

 During iteration $i$ of Algorithm \ref{alg:dragon}, we sample an image $I^{k}$ from a randomly chosen elevation $k$, where $N < k \leq i$, along with its corresponding predicted image $\hat{I}^{k}$, and a predicted image $\hat{I}^{k+1}$ from the previous lower elevation. Our loss function is:
\begin{align}
L_{\name{}}(I^{k}, \hat{I}^{k}, \hat{I}^{k+1}) &= L_{3DGS}(I^{k}, \hat{I}^{k}) \nonumber \\
&\quad + \lambda_{DS} L_\text{DS}(I^{k},\hat{I}^{k+1}) \nonumber \\ 
&\quad + \lambda_{CLIP} L_\text{CLIP}(I^{k},\hat{I}^{k+1}) 
\label{eq:dragon_loss}
\end{align}

where $L_{DS}(\cdot, \cdot)$ and $L_\text{CLIP}(\cdot, \cdot)$ are distances computed using DreamSim and CLIP, and $\lambda_{DS}$ and $\lambda_{CLIP}$ are tradeoff hyperparameters.

%% file: figures_tex/reg_ours.tex
\begin{figure*}[t!]
\centering
\includegraphics[width=0.99\textwidth]{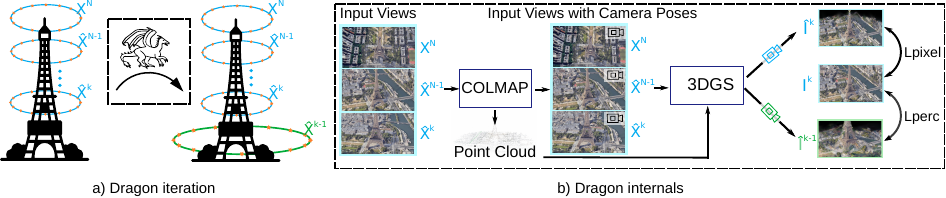}
\caption{ \textbf{Overview of \name{}'s iterative pipeline.} (Left) We iteratively render viewpoints starting from aerial elevations towards ground. Given footage from elevations $X^N, \cdots X^k$, we generate views for elevation $X^{k-1}$. (Right) We feed the accumulated views into COLMAP to obtain the corresponding camera poses and point cloud, which are then passed to the 3DGS model. The model is trained using pixel-wise and perception-wise losses (Equation \ref{eq:dragon_loss}). Blue color denotes training images, while green color represents images rendered for the next elevation.}
\label{fig:reg_ours}
\end{figure*}

%% file: tables/tab_notation.tex
\begin{table}[t!]
\caption{\textbf{Overview of notation used in \name{}.}}
\centering
\begin{tabular}{|c|m{6cm}|}
\hline
Symbol & Description \\
\hline
$X_\text{train}$ & Train images \\
$X_\text{test}$ & Test images \\
$X^{i}$ & Images from elevation level $i$ \\
$X^{0}$ & Images from ground level \\
$X^{N}$ & Images from drone level \\
$X^\text{cum}$ & Images from accumulated elevation levels \\ \hline
$I^{i}$ & Image sampled from elevation level $i$ \\
$\hat{I}^{i}$ & Novel image rendered from elevation level $i$ \\ \hline
$C_\text{train}$ & Camera poses for train images \\
$C_\text{test}$ & Camera poses for test images \\
$C^{i}$ & Camera poses for images from elevation level $i$ \\
$C^\text{cum}$ & Camera poses for images from accumulated elevation levels \\ \hline
model-1H & A "1-headed" model trained on ground truth images from a single elevation \\
model-2H & A "2-headed" model trained on ground truth images from both drone and ground elevations \\ \hline
\end{tabular}
\end{table}

%% file: algorithms/baseline_alg.tex
\begin{algorithm}[t!]
\caption{\bsname{}$(\text{model}, X_\text{train}, C_\text{train}, C_\text{test})$}
\begin{algorithmic}[1]
\Statex Given NVS model, train images $X_\text{train}$, 
\Statex train camera poses $C_\text{train}$, test camera poses $C_\text{test}$
\State $\text{model.train}(X_\text{train}, C_\text{train})$ 
\State $X_{\text{test}} = \text{model.test}(C_\text{test})$ 
\State $\textbf{return } X_\text{test}$
\end{algorithmic}
\label{alg:basic}
\end{algorithm}


%% file: algorithms/dragon_alg.tex
\begin{algorithm}[t!]
\caption{\name{}$(\text{model}, X_\text{train},N)$}
\begin{algorithmic}[1]
\Statex Given NVS model, $X_\text{train} = X^{0} \cup X^{N}$,
\Statex number of elevation levels $N$
\State $X^\text{cum} = X^N$
\For{$i = N-1$ to $0$} \textcolor{gray}{\% iterate over elevations}
    \State $C^\text{cum} = \text{COLMAP}(X^\text{cum})$
    \State $\hat{X}^{i-1} = \text{BASIC}(\text{model}, X^\text{cum}, C^\text{cum}, C^{i-1})$
    \State $X^\text{cum} = X^\text{cum} \cup \hat{X}^{i-1}$
\EndFor
\State $C_\text{train} = \text{COLMAP}(X^\text{cum})$
\State $\textbf{return }\text{BASIC}(\text{model}, X_\text{train}, C_\text{train}, C_\text{test})$
\end{algorithmic}
\label{alg:dragon}
\end{algorithm}

%% file: figures_tex/perceptual_loss.tex
\begin{figure*}[ht!]
\centering
\includegraphics[width=0.99\textwidth]{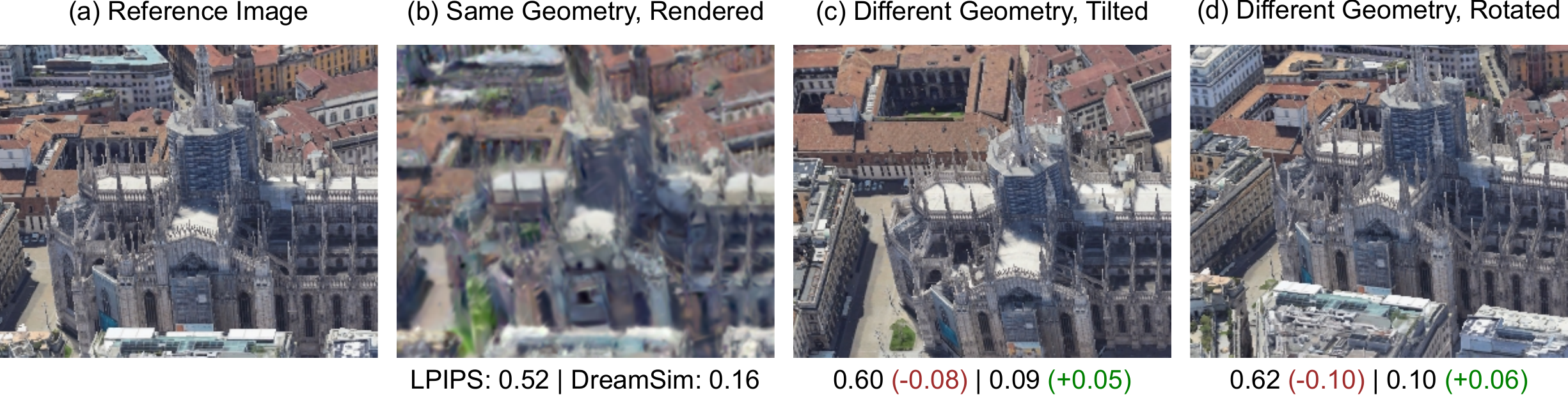}
\caption{\textbf{DreamSim is better than LPIPS at prioritizing perceptual image quality over geometric changes.} While LPIPS score in images with different geometries, (c) tilted and (d) rotated, increased compared to same geometry in image (b), DreamSim score decreased, meaning Dreamsim perceive image (c) and (d) being more similar to the reference image compared to the image (b). This suggests that DreamSim places greater emphasis on perceptual image quality rather than geometric differences (spatial displacement), as long as the image being assessed shares content similarities with the reference image. Perceptual distance measured by DreamSim is more robust to geometry changes than LPIPS metric thus being more suitable for an auxiliary loss function as a regularizer.}
\label{fig:dreamsim}
\end{figure*}

%% file: 5-experiments.tex
\section{Experiments}
\input{tables/tab_quant_eval}
\subsection{Implementation Details}

\subsubsection{Registration}
We use COLMAP for registration because it is the most widely used registration package, and because its GPU-accelerated SIFT~\cite{lowe2004distinctive} feature extraction offers efficient performance. We used an exhaustive matcher between each image pair.

\subsubsection{NVS model}
We adopt 3DGS~\cite{kerbl20233d} as our novel view synthesis method due to its significantly faster training speed, inference time, and reconstruction quality compared to NeRF models~\cite{barron2022mip, muller2022instant, fridovich2022plenoxels}.

We follow the experimental setup in the 3DGS study~\cite{kerbl20233d}. We reset opacity every 3k iterations and apply densification every 100 iterations. We set the spherical harmonic coefficients, which encode the appearance of each Gaussian splat, to 0.0025. We set the opacity, scaling, and rotation parameters to 0.05, 0.005, and 0.001. We use the Adam optimizer~\cite{kingma2014adam} with a learning rate schedule ranging from $1.6 \times 10^{-4}$ to $1.6 \times 10^{-6}$. We set $\lambda_\text{ssim}$ in Eq.~\ref{eq:3dgs_loss} to 0.2. 

\subsubsection{\name{} and Baseline Variants}

Since the basic 3DGS model cannot register both drone and ground images together, we explore two classes of methods: 1-Headed, \ie those trained on images from a single elevation, and 2-Headed, \ie those trained on both drone and ground images. All methods described below use 3DGS, but with different settings of input images and registration schemes. All \name{} methods use the iterative registration scheme outlined in Sec.~\ref{subsec:iterative_scheme}, while Basic and Oracle methods use standard COLMAP registration. 

\textit{1-Headed:} \textbf{Basic} is trained on images from one elevation (either ground or drone) using the loss function in Eq.~\ref{eq:3dgs_loss}.
\noindent \textbf{\name{}-1H Vanilla} is trained on images from one elevation using the loss function in Eq.~\ref{eq:3dgs_loss}. 
\noindent \textbf{\name{}-1H DreamSim} is trained on images from one elevation using the loss function in Eq.~\ref{eq:dragon_loss}.

\textit{2-Headed:}
\noindent \textbf{\name{}-2H Vanilla} is trained with both ground and drone images, using the loss function in Eq.~\ref{eq:3dgs_loss}.
\noindent \textbf{\name{}-2H DreamSim} is trained with both ground and drone images, using the loss function in Eq.~\ref{eq:dragon_loss} with $\lambda_\text{CLIP}$ set to 0. We empirically set $\lambda_\text{DS} = 0.01$ based on the DreamSim distance between images from adjacent elevations. We used the public ensemble DreamSim model using DINO ViT-B/16 \cite{caron2021emerging}, CLIP ViT-B/16 B/32 \cite{radford2021learning}, and OpenCLIP ViT-B/32 \cite{cherti2023reproducible}. 
\textbf{\name{}-2H DreamSim+CLIP} is trained for 25k iterations identical to \name{}-2H DreamSim, and finetuned for another $5k$ iterations with $\lambda_\text{CLIP}=0.01$. We used the public ViT-G/14 OpenCLIP model pretrained on the LAION-2B dataset. 

\textit{Oracle:}
\noindent \textbf{Oracle D\&G} is trained on ground and drone elevation images using Eq.~\ref{eq:3dgs_loss}. \textbf{Oracle All} is trained on images from all elevations (ground, drone, and intermediate levels) using Eq.~\ref{eq:3dgs_loss}. Both oracle methods use ground truth camera poses for all images.

\subsubsection{Training Time}
We train all methods on NVIDIA A100 GPUs. Using a single GPU, training any \name{} method on a building requires roughly 30 minutes, with an inference time of 8 rendered views per second.


\subsection{Evaluation Metrics}\label{subsec:metrics}

\subsubsection{Registration}
To evaluate obtained drone and ground camera poses, we need ground truth camera poses. We establish a pseudo-ground truth registration. To this end, we densely cover the building across elevations, supplementing additional images at intermediate levels and run COLMAP to get dense camera poses. We calculated the percentage of registered matched images and the position and rotation errors for the matched images. Image registration entails a variable coordinate system, influenced by the position of keypoints. Thus, aligning the coordinate system is a prerequisite for accurate error computation against pseudo ground truth ~\cite{mildenhall2019llff, peng2012rasl}. 

\subsubsection{Reconstruction}\label{subsec:metrics}
To quantitatively evaluate reconstruction quality, we use the following metrics: Peak Signal-to-Noise Ratio (PSNR), Structural Similarity (SSIM)~\cite{wang2004image}, and Learned Perceptual Image Patch Similarity (LPIPS)~\cite{zhang2018unreasonable}. 
SSIM assesses structural similarity considering luminance, contrast, and structure, making it less sensitive to color or brightness changes and perception aligned. LPIPS measures the distance between overlapping patches in the input and reconstructed images in a deep feature space. Higher PSNR/SSIM values and lower LPIPS values are preferred as they indicate greater perceptual similarity between images.

\subsection{Registration Results}
\input{tables/tab_big_methods}
We first demonstrate that \name{} enables accurate registration across drone and ground views. Table~\ref{table:registration} presents a quantitative comparison between our registration pipeline and COLMAP applied directly on drone and ground data. 
While COLMAP manages to register drone images, it fails entirely to register ground images due to significant scale variations in levels of detail and field of view. In contrast, our approach achieves near-perfect registration for both drone and ground images. Additionally, our proposed iterative scheme results in significantly lower rotation and position errors.

\subsection{Quantitative Reconstruction Results}

\input{figures_tex/vanilla_vs_dreamsim}
\input{figures_tex/final_render_b1}
\input{figures_tex/final_render_b2}
\input{figures_tex/extrapolated_views}
We categorize the test elevations into three groups: ``drone and ground,'' ``mid elevations'' (intermediate elevations only), and ``all elevations'' (drone, ground, and intermediate). In Table~\ref{table:metrics2}, we present the average PSNR, SSIM, and LPIPS metrics across all nine scenes for these three groups. 
We observe the following from the table:

\subsubsection{1-Headed and 2-Headed Methods}
We find that the Basic with ground images only (1-Headed) is particularly susceptible to occlusions caused by nearby buildings, resulting in significant registration errors, and consequently suboptimal reconstruction performance. Incorporating the DreamSim auxiliary loss yields substantial enhancements across all elevation groups.

Next, we compare our three reconstruction approaches with camera poses obtained using both ground and drone images (2-Headed). 2-Headed approaches outperform 1-Headed, and incorporating auxiliary perceptual losses based on DreamSim and OpenClip yield additional improvements across all metrics for mid elevations.
                        
\subsubsection{Oracle Methods}
Remarkably, our 2-Headed method, which does not assume known camera poses, achieves comparable performance to Oracle Drone \& 
Ground (D\&G), a method that observes ground truth camera poses for drone and ground images. We also report reconstruction results of Oracle All, which has access to both images and camera poses from all elevations. Oracle All surpasses all methods by a significant margin, highlighting the performance gap that subsequent research must address.

\subsection{Qualitative Reconstruction Results}
Fig.~\ref{fig:result-rendering-b1} and Fig.~\ref{fig:result-rendering-b2} provide a qualitative comparison between \name{} and Basic on two scenes: Sydney Opera House and Duomo di Milano. \name{} consistently yields more visually appealing results. While Basic is trained solely on drone images, resulting in slightly better PSNR scores at that specific elevation, \name{} demonstrates superior performance across other elevations, particularly near ground levels. For instance, in the case of the Sydney Opera House, Basic blends the Opera building with the surrounding water area, resulting in an underwater-like rendering for near-ground elevations. This color mixing does not occur with our \name{} approach. Additionally, \name{} produces sharper edges for the opera petals and better preserves the structure of the building.

Similarly, for Duomo di Milano, Basic's performance severely deteriorates near the ground elevation where the rich geometric details of the multiple buildings are completely lost. This is not the case with \name{}, where we can recover the architectural details of the buildings. Even at mid-elevations, \name{} produces sharper details, while Basic hallucinates a black background at the top of the rendered images

We further illustrate the qualitative difference between Dragon-2H DreamSim and Dragon-2H DreamSim+CLIP in Fig.~\ref{fig:result-clip}. When employing DreamSim as our sole perceptual regularizer, occasional high-frequency artifacts are introduced, as exemplified by the Space Needle image. When fine-tuning our model with CLIP, sharper edges are achieved near the center of the image. However, additional haze appears in the lower corners, as observed in the Duomo Di Milano results.

%% file: tables/tab_quant_eval.tex
\begin{table*}[h]
\caption{\textbf{Quantitative registration performance of COLMAP with and without Dragon.} The percentage of registered images averaged over buildings when estimating camera poses separately from the drone and ground level image sets, as well as when estimating camera poses by combining them. It shows that the percentage of registered images higher when using our iterative scheme, and using perceptual regularizer help in registration. Drone and ground imagery sets may not be registered together at once since they have significant scale variations in levels of detail and field of view.}
\label{table:registration}
\centering
\begin{tabular}{|cc|c|c|c|c|c|}
\hline
\multicolumn{2}{|c}{} & \multicolumn{2}{|c}{1-HEADED} & \multicolumn{3}{|c|}{2-HEADED} \\ \cline{3-7}
\multicolumn{2}{|c}{} & \multicolumn{1}{|c|}{COLMAP} & \multicolumn{1}{c|}{COLMAP} & \multicolumn{1}{c|}{COLMAP} & \multicolumn{1}{c|}{\begin{tabular}[c]{@{}c@{}}DRAGON\\ vanilla\end{tabular}} & \multicolumn{1}{c|}{\begin{tabular}[c]{@{}c@{}}DRAGON\\ DreamSim\end{tabular}} \\ \cline{3-7}
\multicolumn{2}{|c}{} & \multicolumn{1}{|c|}{drone-only} & \multicolumn{1}{c|}{ground-only} & \multicolumn{3}{c|}{drone \& ground} \\ \hline
\multicolumn{2}{|c|}{Matched (\%) $\uparrow$}  & 100.00 & 79.58 & 50.00 & 88.00 & \textbf{97.47} \\ \hline
\multicolumn{1}{|c|}{Errors for matched} & rotation ($^{\circ}$) $\downarrow$ & 0.76/0.68 & 19.13/2.39 & 0.69/0.60 & 0.15/0.12 & \textbf{0.12/0.10} \\ \cline{2-2}
\multicolumn{1}{|c|}{(Avg/Std)} & position (m) $\downarrow$ & 0.02/0.02 & 0.50/0.05 & 0.02/0.02 & 0.02/0.02 & \textbf{0.01/0.01} \\ \hline
\end{tabular}
\end{table*}

%% file: tables/tab_big_methods.tex
\begin{table*}
\centering
\caption{\textbf{Quantitative reconstruction performance of view synthesis methods.} We report PSNR, SSIM, and LPIPS metrics averaged across all nine scenes, for three distinct elevation groups: 'drone and ground' (training data), 'mid elevations' (unseen data), and 'all elevations' (combining all data). Comparison is made between 1-headed methods, trained solely on images from one elevation, and 2-headed methods, trained using both ground and drone images. The Oracle methods present the upper bounds of performance.}
\label{table:metrics2}
\begin{tabular}{|m{4.2cm}|*{3}{c|}*{3}{c|}*{3}{c|}}
\hline
 &\multicolumn{3}{c|}{ground and drone} &\multicolumn{3}{c|}{mid elevations} &\multicolumn{3}{c|}{all elevations} \\ \hline
 & PSNR$\uparrow$ & SSIM$\uparrow$ & LPIPS$\downarrow$ & PSNR$\uparrow$ & SSIM$\uparrow$ & LPIPS$\downarrow$ & PSNR$\uparrow$ & SSIM$\uparrow$ & LPIPS$\downarrow$ \\
\hline

\multicolumn{10}{|c|}{2-HEADED METHODS} \\ \hline
\name{}-2H Vanilla   & 25.68 & \textbf{0.85} & \textbf{0.17} & 17.38 & 0.54 & 0.40 & 20.70 & 0.66 & 0.31\\ \hdashline
\name{}-2H DreamSim  & \textbf{25.77} & \textbf{0.85} & \textbf{0.17} & \textbf{18.11} & \textbf{0.57} & \textbf{0.37} & \textbf{21.17} & \textbf{0.68} & \textbf{0.29}\\ \hdashline
\name{}-2H DreamSim+CLIP & 25.76 & \textbf{0.85} & \textbf{0.17} & 18.03 & \textbf{0.57} & \textbf{0.37} & 21.12 & \textbf{0.68} & \textbf{0.29}\\ \hline 

\multicolumn{10}{|c|}{1-HEADED METHODS (GROUND)} \\ \hline
Basic               & 15.97 & 0.47 & 0.45 & 9.35 & 0.24 & 0.65 & 12.00 & 0.33 & 0.57\\ \hdashline
\name{}-1H Vanilla   & 17.55 & 0.54 & 0.41 & 9.02 & 0.26 & 0.66 & 12.43 & 0.37 & 0.56\\ \hdashline
\name{}-1H DreamSim  & \textbf{18.60} & \textbf{0.55} & \textbf{0.38} & \textbf{11.72} & \textbf{0.31} & \textbf{0.57} & \textbf{14.47} & \textbf{0.40} & \textbf{0.49}\\ \hline

\multicolumn{10}{|c|}{1-HEADED METHODS (DRONE)} \\ \hline
Basic               & \textbf{17.07} & \textbf{0.55} & 0.37 & 17.25 & \textbf{0.52} & 0.41 & 17.17 & \textbf{0.53} & 0.39\\ \hdashline
\name{}-1H Vanilla   & 17.06 & \textbf{0.55} & 0.37 & 17.30 & \textbf{0.52} & 0.41 & \textbf{17.21} & \textbf{0.53} & 0.39\\ \hdashline
\name{}-1H DreamSim  & \textbf{17.07} & \textbf{0.55} & \textbf{0.38} & \textbf{17.31} & 0.51 & \textbf{0.42} & \textbf{17.21} & \textbf{0.53} & \textbf{0.40}\\ \hline

\multicolumn{10}{|c|}{1-HEADED VS 2-HEADED} \\ \hline
Basic (ground)        & 15.97 & 0.47 & 0.45 & 9.35 & 0.24 & 0.65 & 12.00 & 0.33 & 0.57\\ \hdashline
Basic (drone)         & 17.07 & 0.55 & 0.37 & 17.25 & 0.52 & 0.41 & 17.17 & 0.53 & 0.39\\ \hdashline
\name{}-2H Vanilla   & 25.68 & \textbf{0.85} & \textbf{0.17} & 17.38 & 0.54 & 0.40 & 20.70 & 0.66 & 0.31\\ \hdashline
\name{}-2H DreamSim  & \textbf{25.77} & \textbf{0.85} & \textbf{0.17} & \textbf{18.11} & \textbf{0.57} & \textbf{0.37} & \textbf{21.17} & \textbf{0.68} & \textbf{0.29}\\ \hdashline
\name{}-2H DreamSim+CLIP & 25.76 & \textbf{0.85} & \textbf{0.17} & 18.03 & \textbf{0.57} & \textbf{0.37} & 21.12 & \textbf{0.68} & \textbf{0.29}\\ \hline 

\multicolumn{10}{|c|}{ORACLE METHODS} \\ \hline
Oracle D$\&$G     & \textbf{26.71} & \textbf{0.88} & \textbf{0.15} & 17.87 & 0.59 & 0.37 & 21.41 & 0.70 & 0.28\\ \hdashline
Oracle All        & 25.04 & 0.83 & 0.18 & \textbf{25.31} & \textbf{0.84} & \textbf{0.18} & \textbf{25.20} & \textbf{0.84} & \textbf{0.18}\\ \hline

\end{tabular}
\end{table*}

%% file: figures_tex/vanilla_vs_dreamsim.tex
\begin{figure}[t!]
\centering
\includegraphics[width=\linewidth]{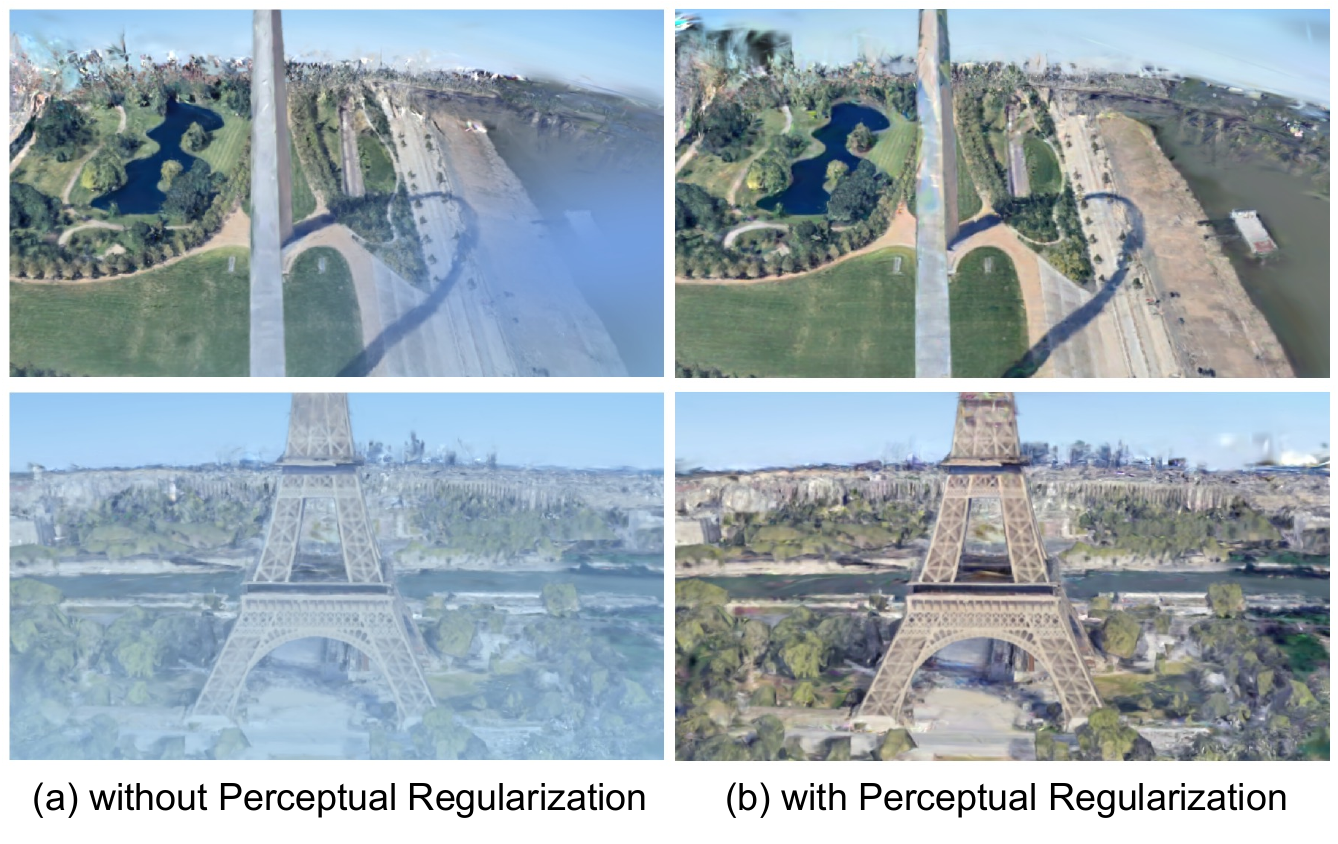}
\caption{\textbf{Perceptual regularization reduces prominent extrapolation artifacts.} Shown here are rendering results for the The GateWay Arch and Eiffel Tower at an intermediate altitude. (a) When training 3DGS on drone and ground image sets without additional regularization, the rendering at intermediate altitude exhibits haziness and artifacts throughout the image. (b) The incorporation of perceptual regularization using DreamSim~\cite{fu2023dreamsim} results in a cleaner rendering.}
\label{fig:vanilla}
\end{figure}

%% file: figures_tex/final_render_b1.tex
\begin{figure*}[t!]
\centering
\includegraphics[width=0.95\textwidth]{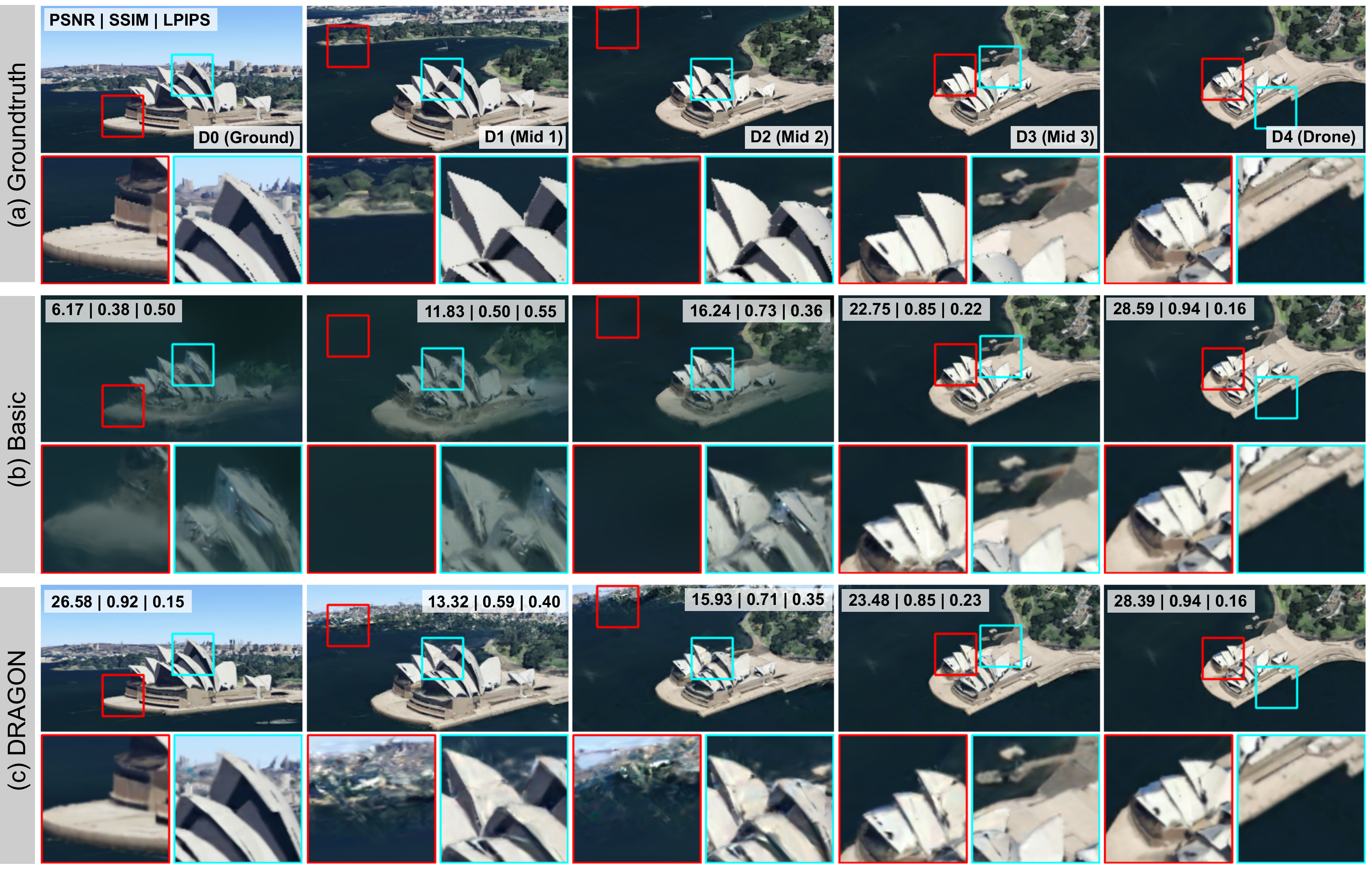}
\caption{ \textbf{Rendering results for the Sydney Opera House.} We present ground truth images (a, top row), along with renderings produced by Basic (b, middle row), and \name{} (c, bottom row). Each column corresponds to a different elevation, from left to right: Ground, Mid 1, Mid 2, Mid 3, and Drone. We also provide metrics (PSNR, SSIM, and LPIPS) directly on the images. The green and cyan boxes below the full images are zoomed in patches. In general, \name{} outperforms Basic qualitatively, particularly for lower elevations. For Mid 2, the image rendered by Basic exhibits higher PSNR and SSIM scores compared to the image rendered by \name{}. But upon closer visual inspection, it is clear that \name{} achieves a more accurate reconstruction of the opera house's true appearance and structure, despite some hallucinatory effects at the top of the image, which may have impacted the quantitative metrics negatively.}
\label{fig:result-rendering-b1}
\end{figure*}

%% file: figures_tex/final_render_b2.tex
\begin{figure*}[t!]
\centering
\includegraphics[width=0.95\textwidth]{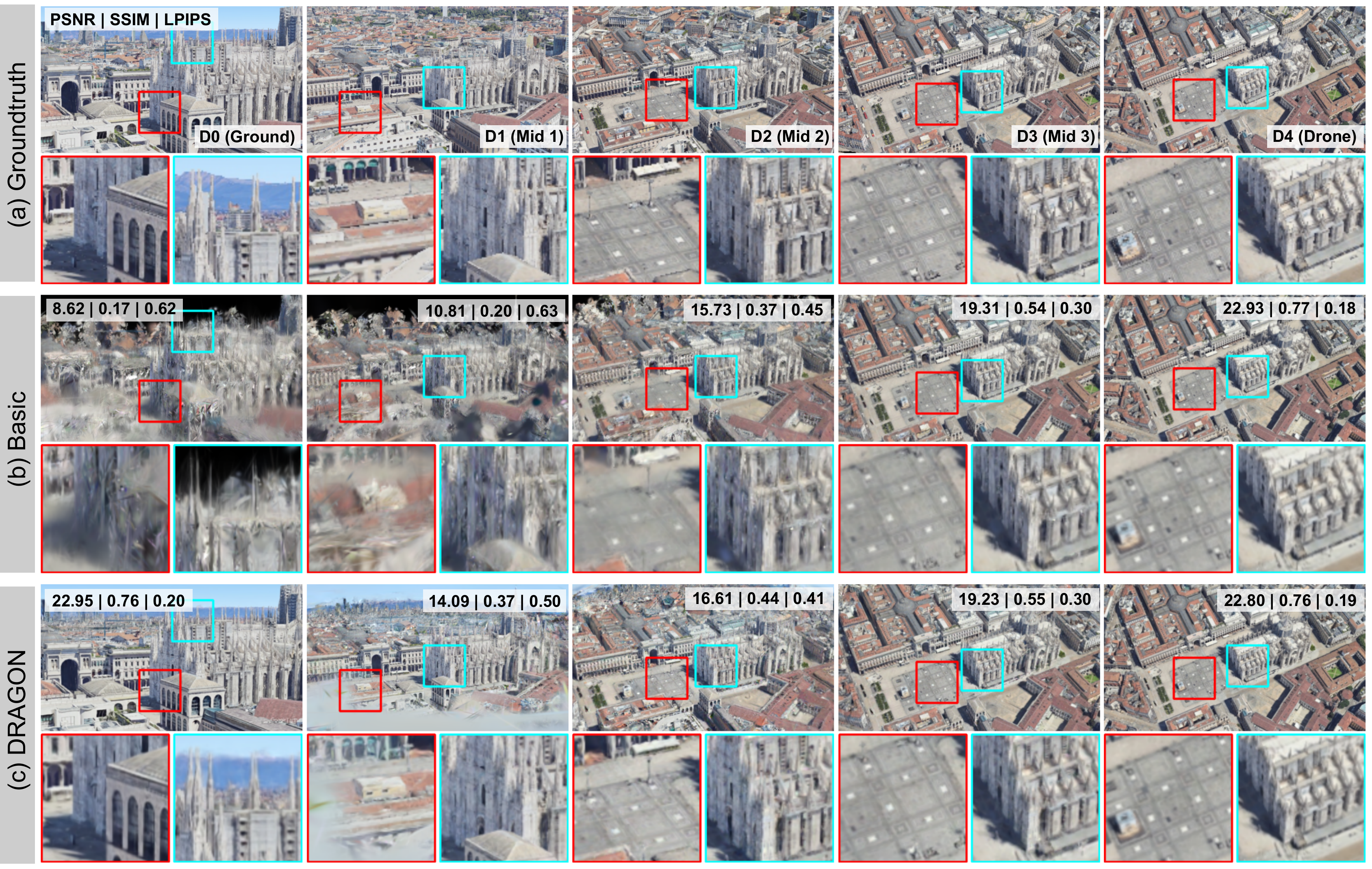}
\caption{\textbf{Rendering results for Duomo di Milano.}  We present ground truth images (a, top row), along with renderings produced by Basic (b, middle row), and \name{} (c, bottom row). Each column corresponds to a different elevation, from left to right: Ground, Mid 1, Mid 2, Mid 3, and Drone. We also provide metrics (PSNR, SSIM, and LPIPS) directly on the images. In general, \name{} outperforms Basic qualitatively, particularly for lower elevations. The green and cyan boxes below the full images are zoomed in patches. For Mid 3, Basic has a higher PSNR score than \name{}, but the latter produces images with more distinct details, particularly in the floor patterns and architecture.}
\label{fig:result-rendering-b2}
\end{figure*}

%% file: figures_tex/extrapolated_views.tex
\begin{figure*}[t!]
\centering
\includegraphics[width=0.9\textwidth]{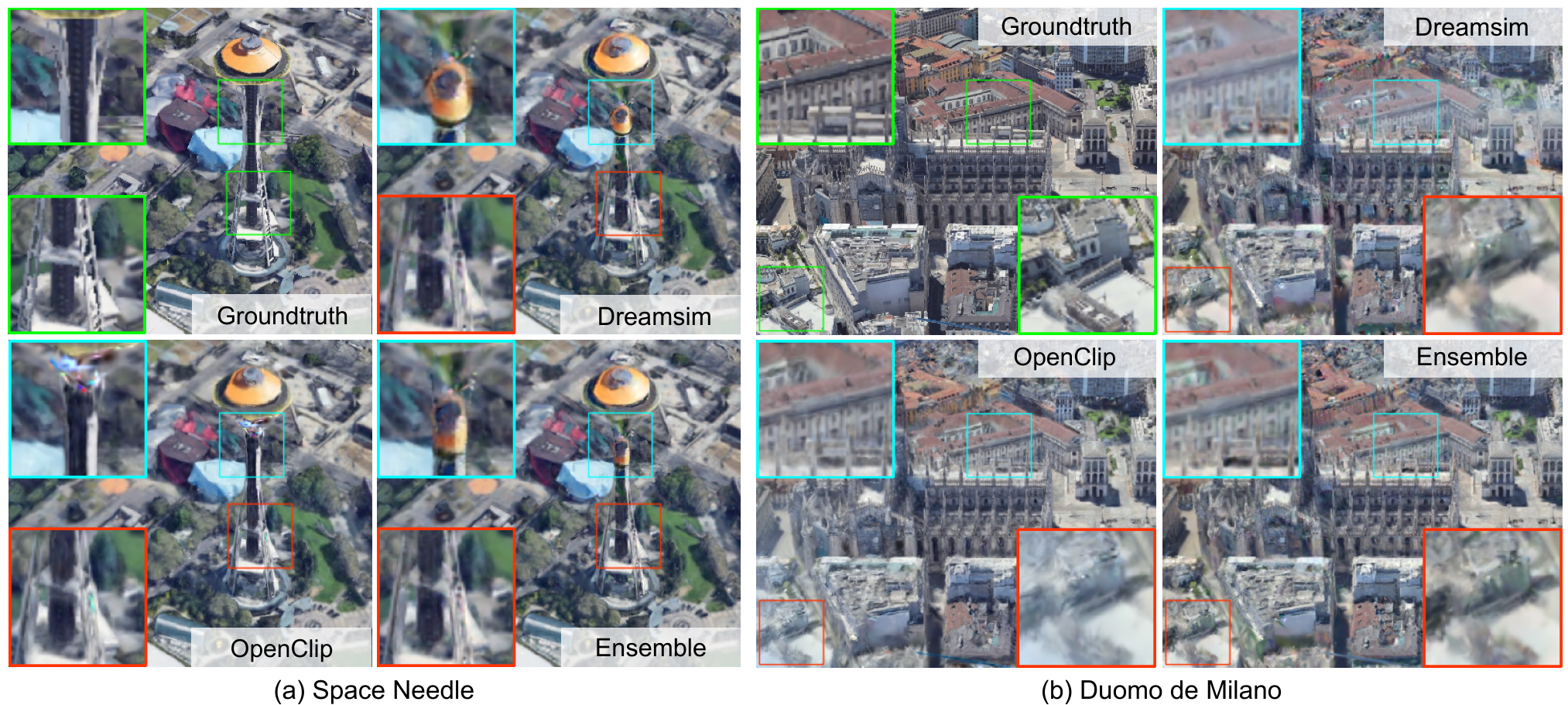}
\caption{ \textbf{Qualitative comparison of \name{} when using DreamSim and CLIP regularizers for Space Needle and Duomo di Milano}. Left: DreamSim introduces artifacts, such as the duplication of the head of the Space Needle (blue box), while CLIP maintains semantic consistency without introducing any additional artifacts. Right: CLIP yields sharper edges of the buildings in the center (blue box), but it introduces additional haze in the lower corner of the image (red box). Conversely, the reconstruction in the corner area with DreamSim does not exhibit this haze.}
\label{fig:result-clip}
\end{figure*}

%% file: 6-conclusion.tex
\section{Discussion and Conclusion}
Results first demonstrate that \name{} offers a significant advantage in terms of registering drone and ground footage together at once and retrieving accurate camera parameters. In contrast, running a registration package like COLMAP on its own essentially fails in matching features across sets with significant disparities. Rendering results compared to various baseline algorithms show that \name{} also provides benefits in terms of rendering quality. In particular, even though we estimated camera poses without any additional information, we achieve comparable or better rendering quality compared to Oracle D$\&$G, which assumes perfect registration.

We observe that perceptual regularization is a key step in improving rendering quality. Without regularization, 3DGS often introduces dramatic artifacts into extrapolated images, such as sky-like features (see Fig.~\ref{fig:vanilla}) and sharp floaters due to sharp transitions in viewpoints from drone to ground or vice versa. While quantitative metrics do not suggest that combining CLIP and DreamSim losses offers an advantage to using DreamSim alone, we observe in the qualitative results (Fig.~\ref{fig:result-clip}) that CLIP and DreamSim capture mutually exclusive visual phenomenon. For example, CLIP seems to better focus details on sharper edges of the buildings near the image center, but it introduces additional haze in the lower image corners. Further human perceptual studies may provide insight into how these perceptual losses interact with one another.

Interestingly, Oracle D$\&$G achieves PSNR/SSIM scores of only roughly 17.87/0.59 on mid elevations, suggesting that even with perfect registration, the standard 3DGS framework exhibits rendering inaccuracies. Some of these shortcomings have been explored in a recent work, VastGaussian~\cite{lin2024vastgaussian}. The focus of our work was not to improve the core 3DGS framework, but to show how an NVS algorithm like 3DGS may be used to perform 3D modeling on large scenes with aerial and ground footage. Our method is agnostic to the exact NVS algorithm used and makes no specific assumptions tied to 3DGS.

While we focused this study on large building reconstruction, the problem of missing viewpoint region chunks (known as the ``missing cone'' problem in tomography~\cite{lim2015comparative}) is a general challenge for NVS methods. The iterative strategy of \name{} may be used for other missing cone scenarios. However, one characteristic specific to building reconstruction that helps the iterative approach is that aerial footage offers a coarse, global context for the structure which may then be refined as the camera lowers in elevation. 

There are several limitations of this work. First, \name{} is a semi-synthetic dataset which allows for ideal and precise acquisition conditions. In contrast, real imagery will have artifacts and imprecise camera positioning. Second, \name{} currently requires a prespecified series of camera poses at intermediate elevations at which to perform novel view synthesis during the iterative process. We currently require these as inputs because it is not trivial to specify them apriori, before a common global coordinate system is established via drone-ground registration. A future direction is to compute these trajectories directly from the drone/ground footage. Third, as shown in Fig.~\ref{fig:result-clip}, our perceptual regularization functions do not completely remove artifacts, and can even inject certain new high-frequency artifacts. This is a well-known shortcoming of using deep neural networks as loss functions for image synthesis~\cite{isola2017image, johnson2016perceptual}.

Finally, measuring perceptual quality of renderings is a difficult challenge on its own. As shown in Fig.~\ref{fig:result-rendering-b1} and Fig.~\ref{fig:result-rendering-b2}, metrics like PSNR/SSIM/LPIPS can be swayed by certain details, particularly those in the background, that may not be meaningful to understand the quality of the building reconstruction alone. One way of accounting for this in the future is to perform human perceptual studies, or design metrics that isolate the structure of interest during evaluation.